\documentclass[letterpaper, 10 pt, conference]{ieeeconf}  

\IEEEoverridecommandlockouts                              
\usepackage{graphics} 
\usepackage{epsfig} 
\usepackage{mathptmx} 
\usepackage{times} 
\usepackage{amsmath} 
\usepackage{amssymb}  
\usepackage{algorithm,algorithmic}
\usepackage{multirow}
\usepackage{epstopdf}
\usepackage{url}
\usepackage[hidelinks]{hyperref}
\usepackage{float}
 \newcommand{\Prob}[1]{{\mathbb P}\left(#1\right)} 

\usepackage{color}

\title{\LARGE \bf
Stopping Rules for Bag-of-Words Image Search and Its Application in Appearance-Based Localization
}

\author{Kiana Hajebi and Hong Zhang
\thanks{K. Hajebi and H. Zhang are with Faculty of Computing Science,
        University of Alberta, Canada,
        {\tt\small [hajebi, hzhang] at ualberta.ca}}%
}

\begin{document}

\maketitle
\thispagestyle{empty}
\pagestyle{empty}

\begin{abstract}

We propose a technique to improve the search efficiency of the bag-of-words method for image retrieval. We introduce a notion of difficulty for the image matching problems and propose methods that reduce the amount of computations required for the feature vector-quantization task in BoW by exploiting the fact that easier queries need less computational resources. Measuring the difficulty of a query and stopping the search accordingly is formulated as a stopping problem. We introduce stopping rules that terminate the image search depending on the difficulty of each query, thereby significantly reducing the computational cost. Our experimental results show the effectiveness of our approach when it is applied to appearance-based localization problem.

\end{abstract}

\section{INTRODUCTION}


Bag-of-Words (BoW) was originally proposed for document retrieval. In recent years, the method has been successfully applied to image retrieval tasks in computer vision community \cite{sivic,nister}. The method is attractive because of its efficient image representation and retrieval. BoW represents an image as a sparse vector of visual words, and thus images can be searched efficiently using an inverted index file system. Other major application areas of the BoW method are appearance-based mobile robot localization and SLAM\footnote{SLAM stands for simultaneous localization and mapping} problems.

The fundamental issue involved with the appearance-based approach to both visual SLAM and global localization is the place recognition. Robot should be able to recognize the places it has visited before to localise itself or refine the map of the environment. This task is performed by matching the current view of the robot to the existing map that contains the images of the previously visited locations. In this paper we consider the problem of appearance-based localization in which the map is known \emph{a priori}.

In large-scale environments maps contain too many images to match. The image search in such a large map is still a challenging and open problem. Matching images by comparing the local features of each image directly to the local features of all other images in the map is not practical. Bag-of-words proposes a more efficient approach; first, rather than matching with a pool of million visual features extracted from many thousands of images, the local features are mapped to a smaller number of vocabulary words that are built in an offline phase. This process is called vector-quantization. Once the visual words are identified in the query image, they are used as indices into the image database, to directly retrieve the images that share the same words.

When the vocabulary is large, the vector-quantization process can be a computationally expensive task in real-time localization.
Considerable research has been done to speed up the search;
some papers \cite{fabmap} do approximate nearest-neighbor search using structures like vocabulary trees \cite{nister}; some methods reduce the number of local image features by selecting only a fraction of features that are highly discriminative \cite{Achar,BoRF}; another group makes use of more compact feature descriptors like \cite{bi-BoW,fabmap}.

In this paper, we first show that some image retrieval tasks are {\em easier} for BoW method. The hardness criteria, defined later, concerns how distinctive the image query is among all images in the dataset. Given this criteria, we show that the BoW search can be terminated earlier for easier queries. This means, in such queries, mapping only a portion of features can be sufficient to yield a relatively good result. The stopping rule saves considerable amount of computational resources.

The intuition behind this is that when there are many similar images to the query image, in terms of the number of common words they share, the search becomes more difficult as more processing is required to find the closest match candidate.
Whereas when the query image has only a few match candidates, i.e., it shares its visual words with only a few images, the search becomes easy as vector-quantizing only a small number of features is sufficient to find the closest match to the query. By exploiting this fact, we can stop the vector-quantization when the search is easy and the nearest neighbor to the query is easy to find. Our method acts as an approximate image search algorithm. Our experimental results show that the accuracy decreases only slightly while the computational cost decreases dramatically.

Our approach can be best compared with the approach of Cummins and Newman \cite{accel_fabmap} who use concentration inequalities (Bennett's inequality in their case) for early bail-out in multi-hypothesis testing that excludes unlikely location hypotheses from further evaluation. However, we use a different bail-out strategy for the process of vector-quantization.


In the next section, we briefly review the image representation and the inverted-index search algorithm used in BoW framework. This is followed by a review of the localization algorithms that employ BoW for the image search. Our proposed method to improve the efficiency of BoW is described in Section~\ref{sec:method}. Section~\ref{sec:results} presents the experimental results and the evaluation criteria, and the result of our comparisons. Finally, we conclude the paper in Section~\ref{sec:conclud}. 

\section{BACKGROUND}
\label{sec:bakgnd}

\subsection{Bag-of-Words for image retrieval}
\label{sec:bow}

Bag-of-words is a popular model that has been used in image classification, objection recognition, and appearance-based navigation. Because of its simplicity and search efficiency it has been used as a successful method in Web search engines for large-scale image and document retrieval \cite{sivic,nister,Sivic2}. 

Bag-of-words model represents an image by a sparse vector of visual words. Image features, e.g., SIFTs \cite{lowe}, are sampled and clustered (e.g., using k-means) in order to quantize the space into a discrete set of visual words. The centroids of clusters are then considered as visual words which form the visual vocabulary. Once a new image arrives, its local features are extracted and vector-quantized into the visual words. Each word might be weighed by some score which is either the word frequency in the image (i.e., \emph{tf}) or the "term frequency-inverse document frequency" or \emph{tf-idf} \cite{sivic}. A histogram of weighted visual words, which is typically sparse, is then built and used to represent the image.

An inverted index file, used in the BoW framework, is an efficient image search tool in which the visual words are mapped to the database images. Each visual word serves as a table index and points to the indices of the database images in which the word occurs. Since not every image contains every word and also each word does not occur in every image, the retrieval through inverted-index file is fast.


\subsection{Bag-of-Words for Image-based Localization}
\label{sec:relwork}

Bag-of-words model has been extensively used as the basis of the image search in appearance-based localization or SLAM algorithms \cite{fabmap,fabmap2,Achar,Angeli,bi-BoW}.
Cummins and Newman \cite{fabmap,fabmap2} propose a probabilistic framework over the bag-of-words representation of locations, for the appearance-based place recognition. Along with the visual vocabulary they also learn the Chow Liu tree to capture the co-occurrences of visual words. Similarly, Angeli \emph{et al.} \cite{Angeli} develop a probabilistic approach for place recognition in SLAM. They build two visual vocabularies incrementally and use two BoW representations as an input of a Bayesian filtering framework to estimate the likelihood of loop closures.

Assuming each image has hundreds of SIFT features, mapping the features to the visual words, using a linear search method, is computationally expensive and not practical for real-time localization. Researchers have tackled this problem with different approaches that speeds up the search but at the expense of accuracy. A number of papers have employed compact feature descriptors that speeds up the search. G\'{a}lvez-L\'{o}pez and Tard\'{o}s in \cite{bi-BoW} propose to use FAST \cite{FAST} and BREIF \cite{BRIEF} binary features and introduce a BoW model that descritizes a binary space. Similarly, \cite{fabmap,fabmap2} use SURF \cite{Surf} to have a more compact feature descriptor. Another approach is to use approximate nearest neighbor search algorithms, like hierarchical k-means \cite{nister}, KD-trees \cite{fabmap} or graph-based search methods \cite{GNNS} and \cite{SGNNS}, to speed up the quantization process.

Achar \emph{et al.} \cite{Kosecka} and Zhang \cite{BoRF} propose reducing the number of features in each image, thereby reducing (removing) the vector-quantization process. They keep track of the features that are repeatable over time. However, these approaches are more specific to the navigation problems where the data is sequential and there exists considerable overlaps between consecutive images. Our approach is more similar to this group as we also reduce the amount of feature mapping. However rather than only selecting a small set of features, we use all features but we stop the mapping process when necessary. Our approach is also more general as it does not depend on the sequential property of the data.


\section{PROPOSED METHOD}
\label{sec:method}
\begin{figure}[t]
\centering%
\begin{tabular}{cc}
 \includegraphics[width=.21\textwidth]{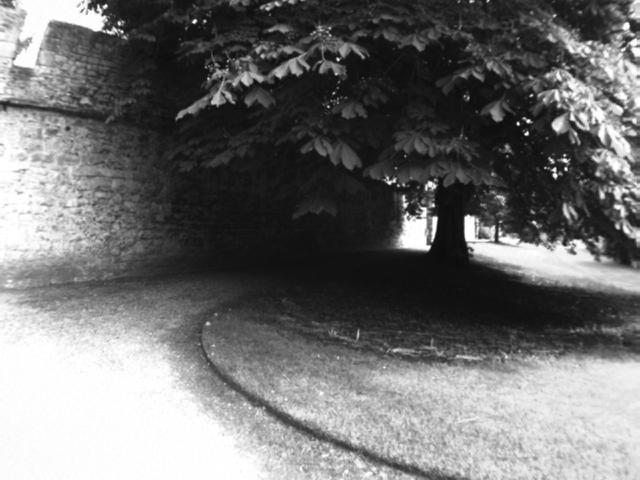} & \includegraphics[width=.21\textwidth]{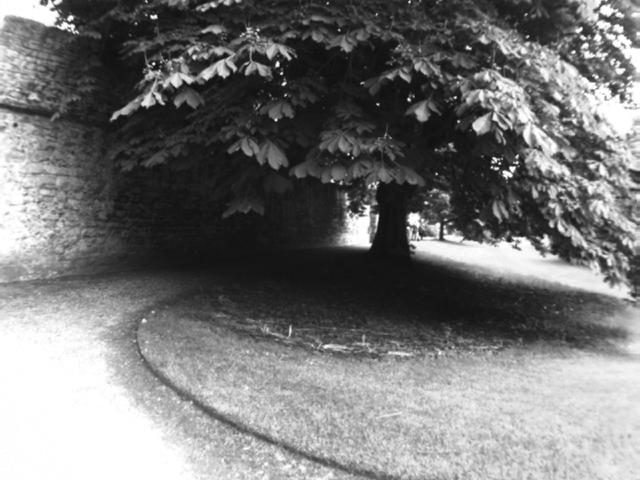} \\
 (a) & (b) \\
 \includegraphics[width=.22\textwidth]{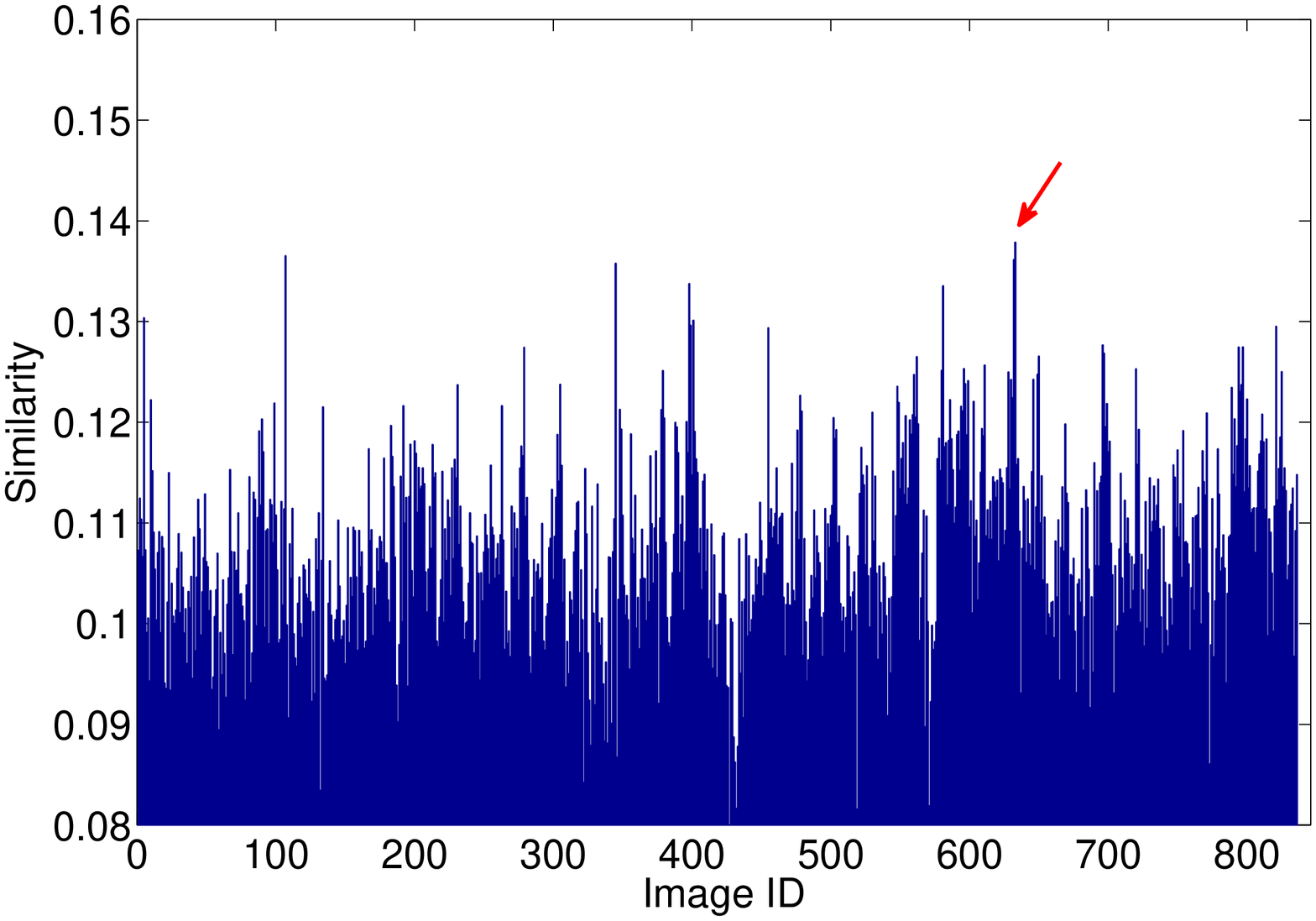} &
 \includegraphics[width=.223\textwidth]{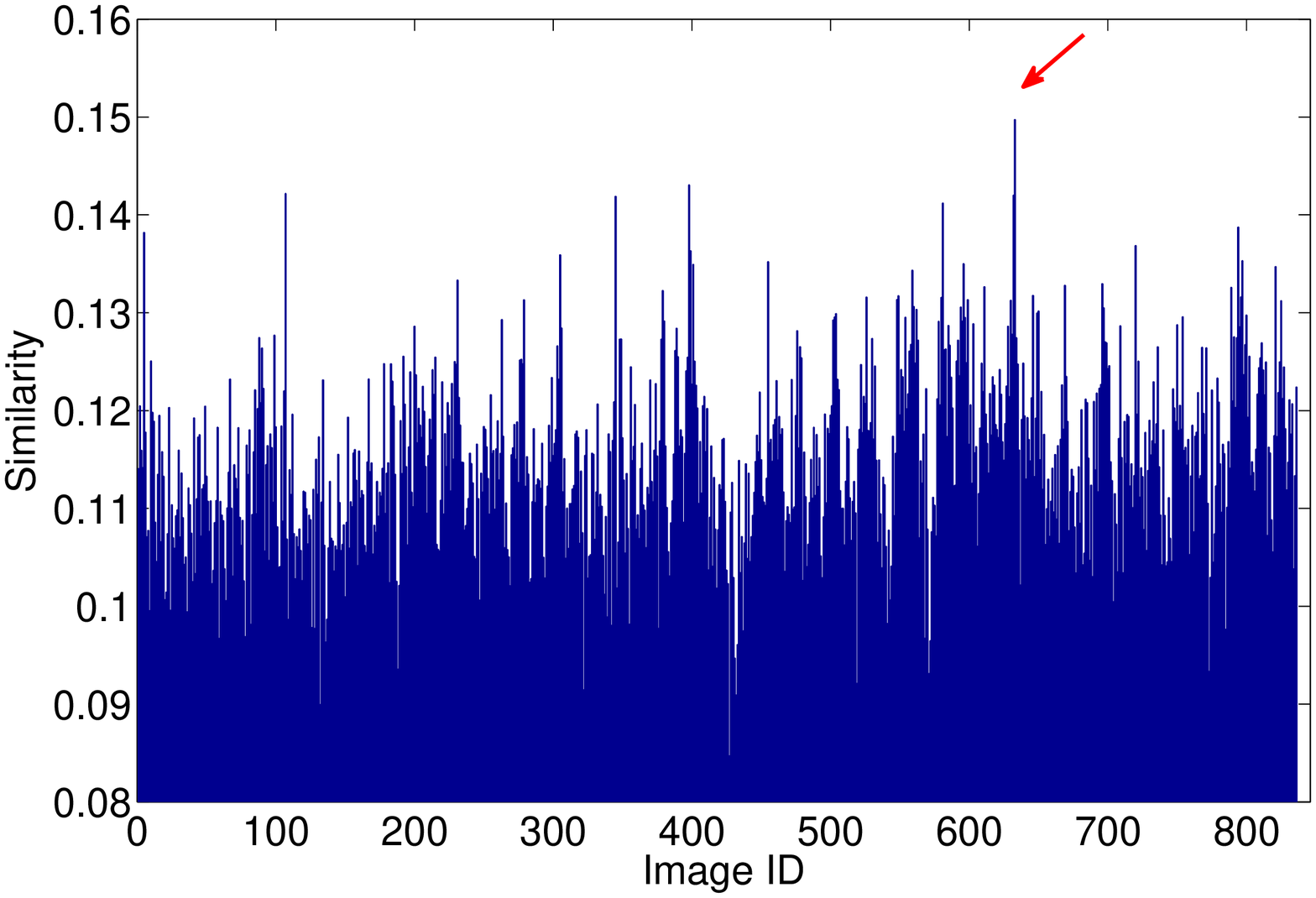}  \\
 (c) & (d) \\
\end{tabular}
\caption{Difficult search. The Query image (a) has been matched to (b). (c) shows the histogram of images' scores after vector-quantizing 89\% of features. The peak of the histogram (shown by the red arrow) does not change until the last histogram (d) when all features have been processed. This means processing the last 11\% of features (or the votes of the last 11\% of words) are not necessary to select the highest scored image.}
\label{diff_search}
\begin{tabular}{cc}
 & \\
 &
\end{tabular}
\begin{tabular}{cc}
 \includegraphics[width=.21\textwidth]{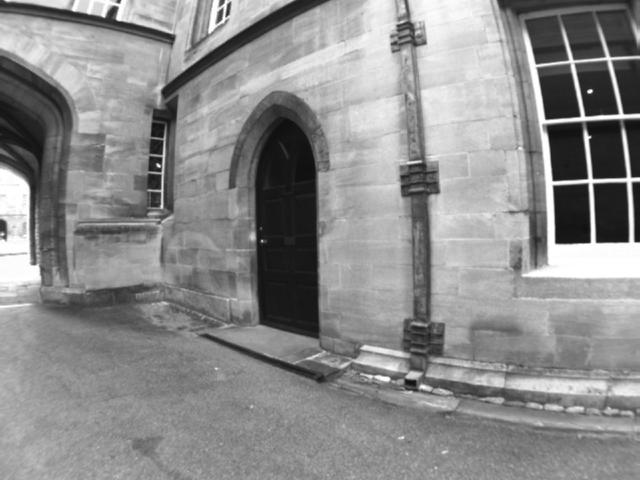} & \includegraphics[width=.21\textwidth]{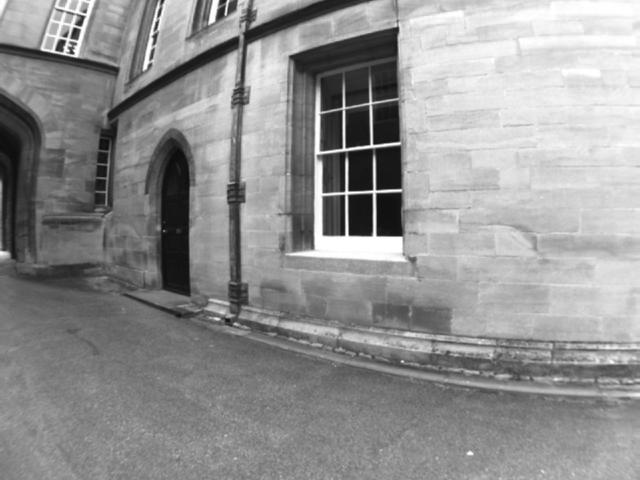} \\
  (e) & (f) \\
 \includegraphics[width=.22\textwidth]{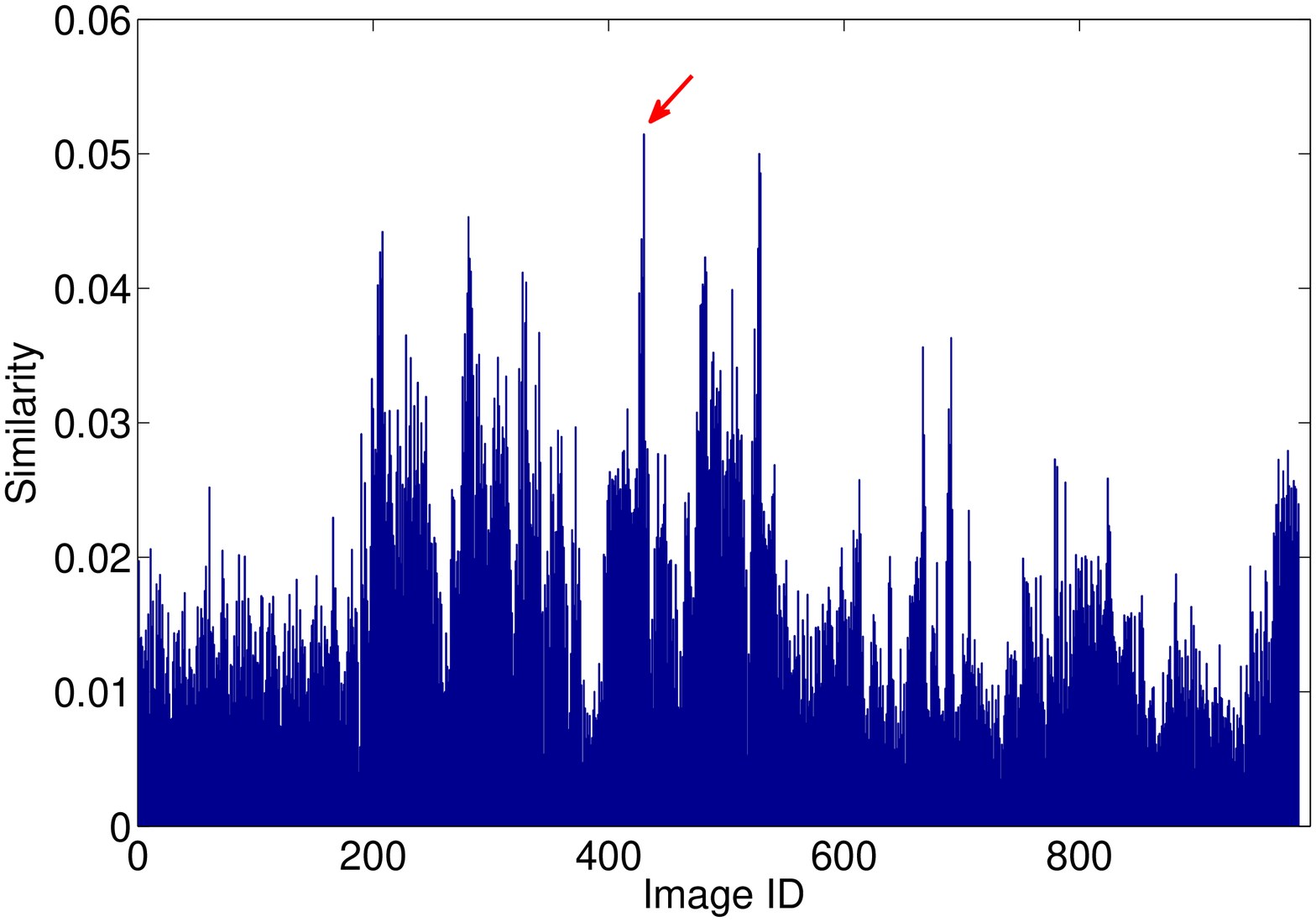} &
 \includegraphics[width=.223\textwidth]{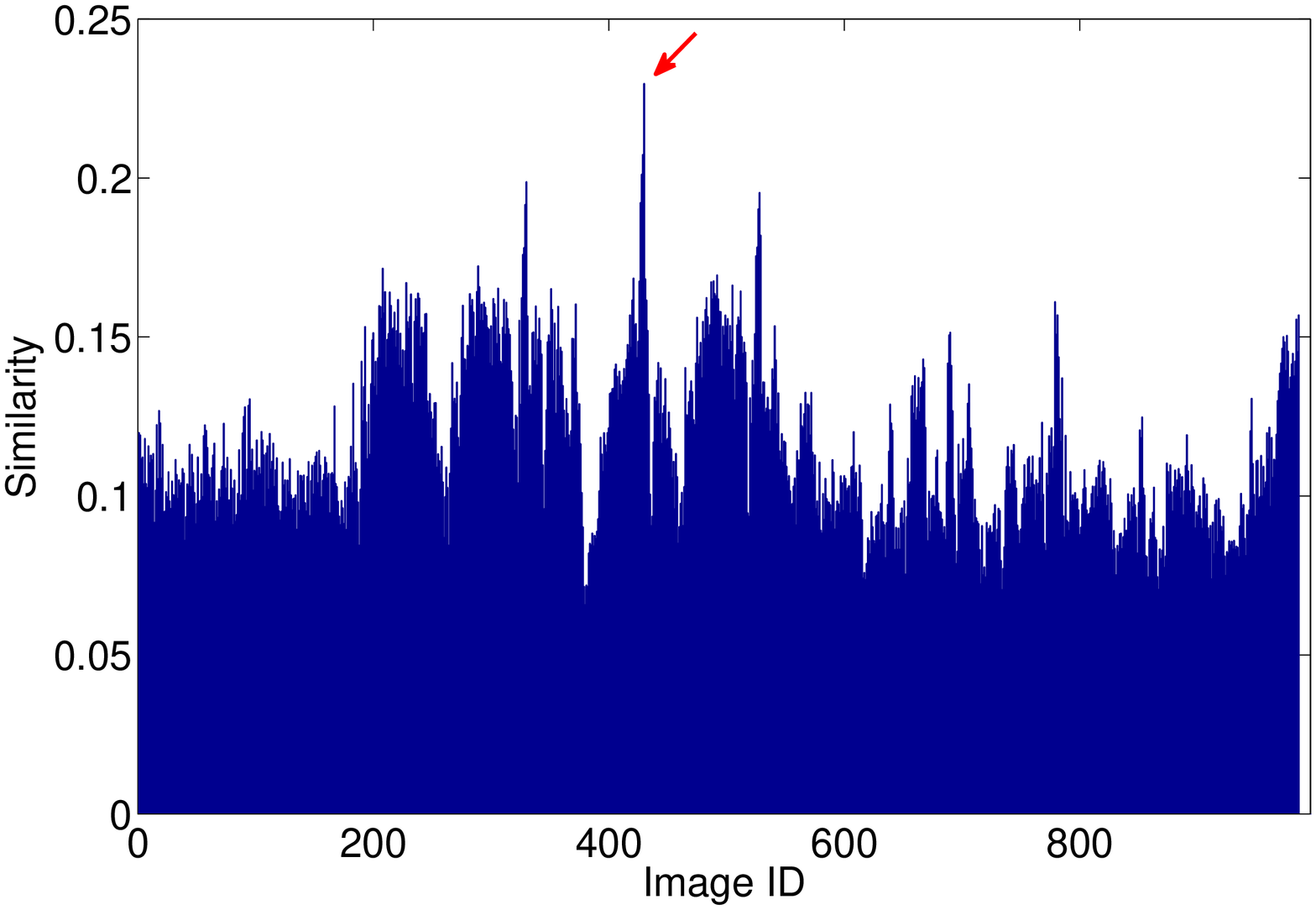} \\
 (g) & (h) \\
\end{tabular}
\caption{Easy search. The Query image (a) has been matched to (b). (c) shows the histogram of images' scores after vector-quantizing 12\% of features. The peak of the histogram (shown by the red arrow) does not change until the last histogram (d) when all features have been processed. This means processing the last 88\% of features (or the votes of the last 88\% of words) are not necessary to select the highest scored image.}
\label{easy_search}
\end{figure}

Vector-quantization (VQ) is an expensive process when the BoW-based image retrieval is performed in large-scale environments, in which hundreds of features extracted from an image need to be matched against hundreds of thousands of visual words. The question is if we really need to vector-quantize all features? Depending on the difficulty level of the search, the number of features to be converted to visual words may vary. We call a search difficult when there are many similar images to the query and thus finding the nearest neighbor among all those candidates requires more computations
\footnote{The similarity between two images can be simply defined as the number of visual words they share. Other similarity measures like tf\_idf \cite{sivic} have been used as well.}.
Whereas in an easy search, the query image is similar to only a few images in the database and it can find its true match after processing only a small percentage of features. Figure~\ref{diff_search} and Figure~\ref{easy_search} show examples of easy and difficult searches. The histograms show the similarity of each database image to the query based on the \emph{tf-idf} score. The difficult search needs to process at least $89\%$ of features to find the closest match to the query (indicated by the peak of the histogram), however the easy search can stop the search after processing only $12\%$ of features as the peak does not change until the end of the search. Comparing the distance between the peak and average of the histograms, it can be seen that in difficult search this distance is smaller than that in easy search, which is expected.

Initially each image starts with no vote. In original BoW, image features are converted to visual words one by one and the histogram (of images' scores) is built incrementally. Each bin of the histogram corresponds to one of the database images and indicates the score of that image. Each visual word will cast a distance-weighted vote for multiple images, i.e., histogram bins. This process continues until all words cast their votes and then the peak of the histogram (the bin with the highest score) determines the nearest neighbor to the query.\footnote{Usually, rather than returning the peak as the true match, a number of highest scored images are selected as match candidates and then the true match is selected after further processing like geometric verification among those candidates. Here for the simplicity, we consider the peak of the histogram as the true match.}

\begin{table}[th]
\caption{The proposed voting scheme and stopping rules.}
\label{alg:mapping}
\begin{center}
\begin{tabular}{c}
\framebox{\parbox{3.2in}{ 
\begin{algorithmic}
\STATE \textbf{Function} inverted-index
\STATE \textbf{Input}: $im_q$: current view of the robot, $N_{im}$: the number of images in the database, $T$: stopping threshold for mapping.
\STATE Let $h$ be a histogram of length $N_{im}$, with all bins initialized to zero
\STATE $f_q$ = extract-feature$(im_q)$.
\STATE $r_q$ = random-permute$(f_q)$.
\STATE $i=0$.
\WHILE {stop-time$(h,T) == $FALSE}
    \STATE $i = i+1$.
    \STATE $w$ = vector-quantize$(r_q(i))$.
    \STATE Find the list of images containing word $w$, $L$ = inverted-index$(w)$.
    \FOR {$j = 1 \dots |L|$}
        \STATE $h(L(j))= h(L(j)) + \mbox{score}(w,L(j))$.
    \ENDFOR
\ENDWHILE
\STATE return $\max(h)$ as the true match
\end{algorithmic}
}} \\ \\

\framebox{\parbox{3.2in}{ 
\begin{algorithmic}
\STATE \textbf{Function} stop-time
\STATE \textbf{Input}: $h$: histogram of images' scores, $T$: stopping threshold.
\IF {$|\max(h)-\mbox{mean}(h)| > T$}
\STATE return TRUE;
\ELSE
\STATE return FALSE;
\ENDIF
\end{algorithmic}
}}

\end{tabular}
\end{center}
\end{table}

Our method builds upon the voting scheme of BoW; See Table~\ref{alg:mapping}. The features are selected randomly to be vector-quantized and are used to score the images (performed by \texttt{\small extract-feature}, \texttt{\small random-permute}, and \texttt{\small vector-quantize} functions in Table~\ref{alg:mapping}).
In an easy search, as the query has many common words with its nearest neighbor and not with other images, after processing only some small percentage of the features the score of the true matching image becomes significantly larger than the mean of the histogram.
However, when the query shares its words with many images, i.e., a difficult search, a higher percentage of the words need to cast their vote to find the true match.
We stop the feature mapping once the distance between the peak of the histogram to the average of the other bins is greater than some threshold (\texttt{\small stop-time} function). Other stopping rules might be defined, which are described in Section~\ref{sec:rules}.

Our method is different from the naive approach of stopping after quantizing a fixed percentage of the features. Based on our stopping criterion when the search is easy, the VQ stops sooner, otherwise it stops after processing more features. With the naive approach, a search is forced to stop even if more processing is required to find the nearest neighbor.

The problem that we study in this paper can be formulated as a stopping problem \cite{Shiryaev}. The stopping problem is a decision making problem where at each round, the decision maker observes an input and a (possibly noisy) reward and decides whether he wants to see the next input or not. The objective is to maximize the expected reward when he decides to {\em stop}. The stopping problem is studied in various settings in statistics, decision theory, and economics.

More specifically, we can formulate our problem as finding the best option from a pool of options based on some measurements. Assume a number of distributions are given, from which we want to find the one with the highest mean. At each round, we observe a new sample from each distribution. As sampling can be expensive, the objective is to stop sampling and find the distribution with the highest mean as quick as possible. The common approach is to employ certain concentration inequalities (such as Hoeffding's Inequality \cite{Hoeffding}) to construct confidence bands around the empirical mean of each distribution after a number of observations. When the approximations are accurate enough and the decision maker has enough confidence, he stops and chooses the distribution with the highest empirical mean \cite{Hoeffding_races}.

To illustrate the ideas, assume we are given two distributions, $p_1$ and $p_2$, and are asked to find the one with the higher mean. Let the expected values of these distributions be $\mu_1$ and $\mu_2$, respectively. Assume that $\mu_1>\mu_2$. Let $x_s\in[0,1], 1\le s\le t$ be samples from $p_1$ and $y_s\in[0,1], 1\le s\le t$ be samples from $p_2$. Define the empirical means by $\overline{X}_t = \frac{1}{t}\sum_{s=1}^t x_s$ and $\overline{Y}_t = \frac{1}{t}\sum_{s=1}^t y_s$ and the empirical gap by $g_t = \overline{X}_t - \overline{Y}_t$. Without loss of generality, assume that $g_t > 0$. From Hoeffding's Inequality, we get that

\[
\Prob{\overline{X}_t - \mu_1 \le g_t/2} \ge 1 - e^{-g_t^2 t/2}\,,
\]
and
\[
\Prob{\overline{Y}_t - \mu_2 \ge -g_t/2} \ge 1 - e^{-g_t^2 t/2}\;.
\]
Thus, with probability at least $1-2e^{-g_t^2 t/2}$, $\overline{X}_t - \mu_1 \le g_t/2$ and $\overline{Y}_t - \mu_2 \ge -g_t/2$, which implies that
\[
\mu_1 > \mu_2\;.
\]
Thus, we have correctly identified the distribution with the highest mean (here $p_1$) with probability at least $1-2e^{-g_t^2 t/2}$. If we demand that this probability be at least $1-\delta$ for some $\delta\in (0,1)$, then we need to have
\begin{equation}
\label{eq:num_samples}
g_t\ge \sqrt{\frac{2\log(2/\delta)}{t}}\;.
\end{equation}
Equation~\eqref{eq:num_samples} can be used as a stopping condition to make the right decisions with high probability. We might ask how many samples are needed before Condition~\eqref{eq:num_samples} is satisfied? By applying Hoeffding's Inequality, it is not difficult to see that if the true gap ($g=\mu_1-\mu_2$) is small, then the number of samples $t$ in Condition~\eqref{eq:num_samples} needs to be larger. This implies that the identification is more difficult when there is a small gap between the two distributions.

In our image matching problem, given a query image, we want to find an image that is closest in terms of an approximate cosine distance. The BoW procedure processes the features one by one, increasing the score of each candidate image by some number. We can view the candidate images as different distributions and the new scores as the new samples. At each round, we have a new estimate for the similarity of each candidate image to the query image. The problem is to stop sampling and return the true match as quick as possible.

It might seem natural to use the stopping condition~\eqref{eq:num_samples} for this problem. However, we found out that these theoretical results are very conservative in practice and often require the processing of a large number of features before stopping, so we do not test Hoeffding's inequality experimentally. In the next section, we propose a number of rules that show better performance.





\section{EXPERIMENTAL RESULTS}
\label{sec:results}

In this section, we compare the performance of the original BoW method with the BoW method that uses our stopping rules, when applied to the appearance-based localization problem. We will describe the datasets we used for performance evaluation of both methods, followed by discussion of our experimental results. We also describe the different stopping rules that we used in our experiments.

\subsection{Datasets}

\begin{figure}[th]
\centering%
 \tabcolsep 1pt
\begin{tabular}{ccc}
 \includegraphics[width=.15\textwidth]{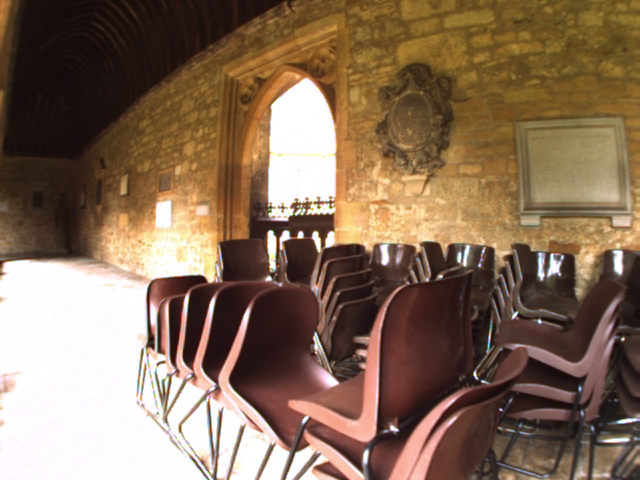} & \includegraphics[width=.15\textwidth]{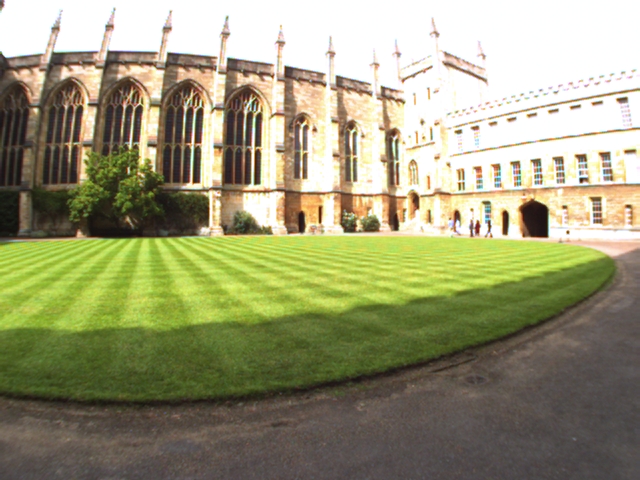} &
 \includegraphics[width=.15\textwidth]{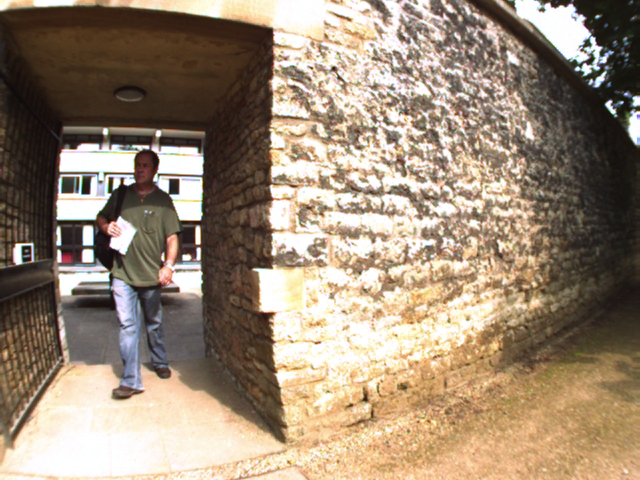} \\

 \includegraphics[width=.15\textwidth]{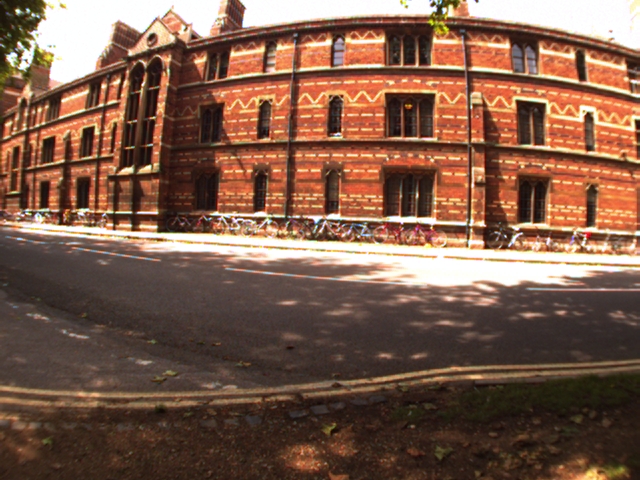} & \includegraphics[width=.15\textwidth]{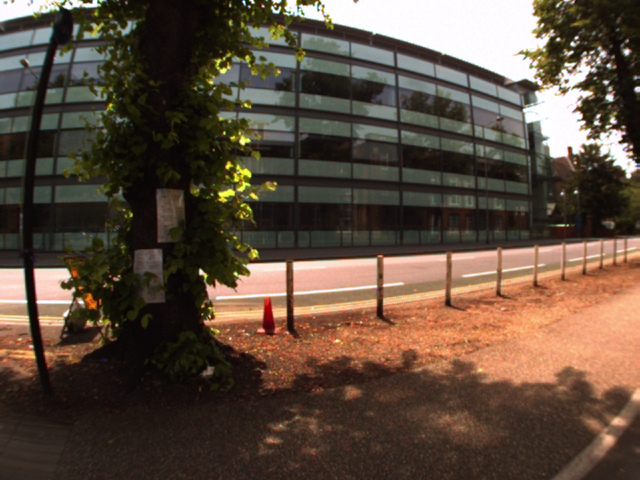} &
 \includegraphics[width=.15\textwidth]{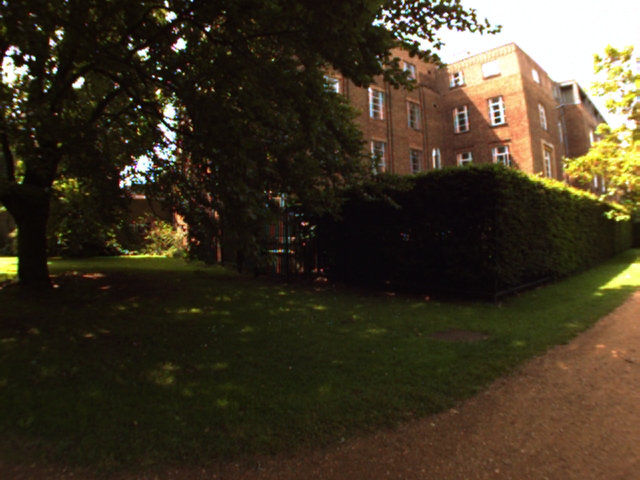}  \\
\end{tabular}
\caption{Images from the New College (top row) and City Center (bottom row) sequences used in the reported experiment.}
\label{fig:datasets}
\end{figure}

We performed our experiments on four datasets. Two datasets have been selected from the Oxford City Center dataset and two from the New College dataset \cite{fabmap} (see Figure~\ref{fig:datasets}). Both have been used as the benchmark for localization and SLAM evaluations. The ground truth data is also available for each of them\footnote{\url{http://www.robots.ox.ac.uk/~mobile/IJRR_2008_Dataset/}}. Each dataset contains two sets of image sequences. One sequence is taken from the right camera and the other from the left camera mounted on the robot. Each sequence of the City Center dataset contains $1237$ images and each of the New College contains $1073$. The resolution of images is $640\times480$. For each dataset we used the first half of the images as training data on which we performed the \emph{k-means} clustering and generated a vocabulary of $5000$ visual words. We used $128$-dimensional SIFT feature descriptors as the input to the clustering. Each image has $\sim 400$ SIFT descriptors on average.

The other half of the sequences have been used as the test data, i.e., query images. Each query is matched to the earlier images in the sequence. For each query there are multiple matches in ground truth. If the match that we find for each query is among those correct(ground truth) matches, we call the match correct otherwise incorrect.


\subsection{Performance Evaluations}

On each dataset we ran the BoW with the original voting scheme (inverted-index) and BoW with our voting scheme that employs stopping rules. 
Each visual word in the query image has been weighed with the \emph{tf-idf} score and the top-$n$ images with highest scores have been returned as match candidates. We set $n$ to 3, 5 and 10 in our experiments. We select the candidates whose similarity to the query is above some threshold.

Our experimental results have been summarized in Table~\ref{exp1:city1} to Table~\ref{exp1:newcol2}. 
The reported recall values are for precisions above 90\%. Top 3, top 5 and top 10 in Tables~\ref{exp1:city1} to \ref{exp1:newcol2} indicate the top-3, -5 and -10 nearest neighbors to the query image that we retrieved. We used different stopping thresholds to generate different recalls. The percentage of the features that we processed and the accuracy of the localization have been computed and compared with the original BoW. All the results have been produced by averaging over $10$ Monte Carlo runs. The experiments show that we have improved the computational cost significantly at the expense of slight reduction in accuracy.


\begin{table}[!th]
\caption{Comparison of our approach to the original BoW on City Center dataset, right-side sequence, 2631 words.}
\label{exp1:city1}
\tabcolsep 4.5pt
\begin{tabular}{|c|c|c||c|c|c|c|}
  \cline{3-7}
  \multicolumn{1}{c}{} &  &    BoW          & \multicolumn{4}{c|} {BoW} \\
  \multicolumn{1}{c}{} &  & (Inv. Indx)     & \multicolumn{4}{c|} {(with our voting scheme)} \\  \hline
  \multirow{3}{*}{Recall}
   & top 3:  & 0.7897   & 0.7861 & 0.7540 & 0.7273 & 0.6934  \\
   & top 5:  & 0.8378   & 0.8342 & 0.8111 & 0.7932 & 0.7647  \\
   & top 10: & 0.9055   & 0.9037 & 0.8806 & 0.8610 & 0.8503  \\
  \hline
  \multicolumn{2}{|c|} {\% Features}    & 1      & 0.8655 & 0.6879 & 0.5911 & 0.4965 \\  \hline
  \multicolumn{2}{|c|} {Threshold}      & -      & 0.25   & 0.2    & 0.18   & 0.16   \\  \hline
\end{tabular}

\begin{tabular}{cc}
 & \\
 \vspace{.2cm}
\end{tabular}

\caption{Comparison of our approach to the original BoW on the City Center dataset, left-side sequence, 5000 words.}
\label{exp1:city2}
\tabcolsep 4.5pt
\begin{tabular}{|c|c|c||c|c|c|c|}
  \cline{3-7}
  \multicolumn{1}{c}{} &  &     BoW      & \multicolumn{4}{c|} {BoW} \\
  \multicolumn{1}{c}{} &  & (Inv. Indx)  & \multicolumn{4}{c|} {(with our voting scheme)} \\  \hline
  \multirow{3}{*}{Recall}
   & top 3:   & 0.8324   & 0.8253 & 0.8164 & 0.7879 & 0.7219  \\
   & top 5:   & 0.8556   & 0.8538 & 0.8485 & 0.8235 & 0.7772  \\
   & top 10:  & 0.8895   & 0.8877 & 0.8895 & 0.8717 & 0.8307  \\
  \hline
  \multicolumn{2}{|c|} {\% Features}   & 1      & 0.8395 & 0.6606 & 0.5001 & 0.3240 \\  \hline
  \multicolumn{2}{|c|} {Threshold}     & -      & 0.28   & 0.20   & 0.15   & 0.10 \\  \hline
\end{tabular}

\begin{tabular}{cc}
 & \\
 \vspace{.2cm}
\end{tabular}

\caption{Comparison of our approach to the original BoW on the New College dataset, right-side sequence, 600 words.}
\label{exp1:newcol1}
\tabcolsep 4.5pt
\begin{tabular}{|c|c|c||c|c|c|c|}
  \cline{3-7}
  \multicolumn{1}{c}{} &  &     BoW      & \multicolumn{4}{c|} {BoW} \\
  \multicolumn{1}{c}{} &  & (Inv. Indx)  & \multicolumn{4}{c|} {(with our voting scheme)} \\  \hline
  \multirow{3}{*}{Recall}
   & top 3:   & 0.9104   & 0.8920 & 0.8668 & 0.8455 & 0.7743  \\
   & top 5:   & 0.9395   & 0.9322 & 0.9211 & 0.8988 & 0.8416  \\
   & top 10:  & 0.9709   & 0.9642 & 0.9613 & 0.9535 & 0.9138  \\
  \hline
  \multicolumn{2}{|c|} {\% Features}   & 1      & 0.7821 & 0.6670 & 0.4994 & 0.3030 \\  \hline
  \multicolumn{2}{|c|} {Threshold}     & -      & 0.12   & 0.10   & 0.08   & 0.06 \\  \hline
\end{tabular}

\begin{tabular}{cc}
 & \\
 \vspace{.2cm}
\end{tabular}

\caption{Comparison of our approach to the original BoW on the New College dataset, left-side sequence, 1395 words.}
\label{exp1:newcol2}
\tabcolsep 4.5pt
\begin{tabular}{|c|c|c||c|c|c|c|}
  \cline{3-7}
  \multicolumn{1}{c}{} &  &     BoW      & \multicolumn{4}{c|} {BoW} \\
  \multicolumn{1}{c}{} &  & (Inv. Indx)  & \multicolumn{4}{c|} {(with our voting scheme)} \\  \hline
  \multirow{3}{*}{Recall}
   & top 3:   & 0.9487   & 0.9397 & 0.9299 & 0.8778 & 0.8150  \\
   & top 5:   & 0.9780   & 0.9707 & 0.9544 & 0.9356 & 0.8818  \\
   & top 10:  & 0.9902   & 0.9894 & 0.9804 & 0.9658 & 0.9487  \\
  \hline
  \multicolumn{2}{|c|} {\% Features}   & 1      & 0.8567 & 0.7374 & 0.5517 & 0.3211 \\  \hline
  \multicolumn{2}{|c|} {Threshold}     & -      & 0.25   & 0.20   & 0.15   & 0.10 \\  \hline
\end{tabular}
\end{table}

\subsection{Different Stopping Rules}
\label{sec:rules}
\begin{figure}[t]
\centering%
\begin{tabular}{c}
 \includegraphics[width=.40\textwidth]{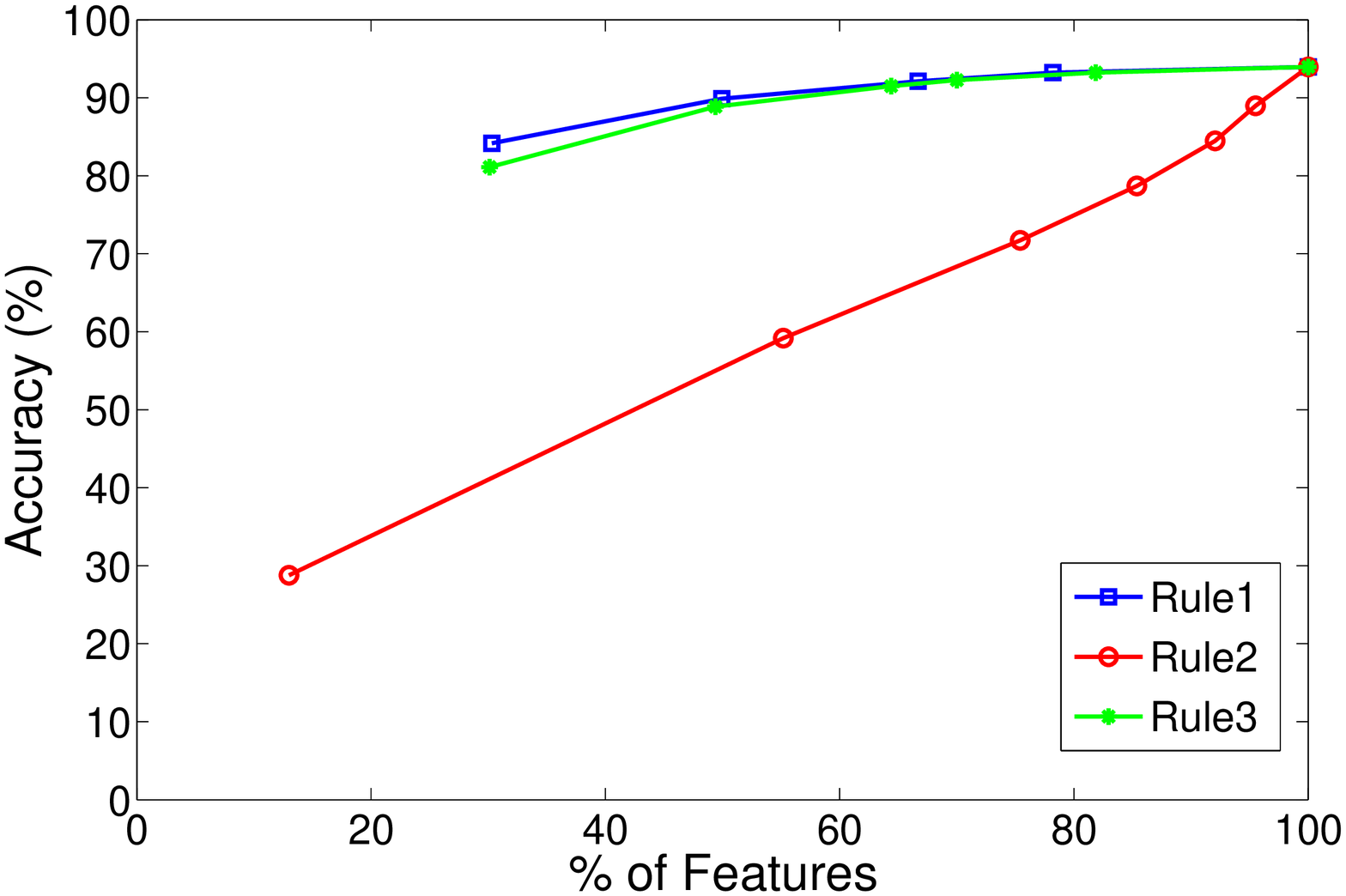}
\end{tabular}
\caption{Accuracy (recall) vs. \%  of features used, New College dataset}
\label{fig:rules}

\begin{tabular}{c}
 \\
\end{tabular}

\centering%
\begin{tabular}{c}
 \includegraphics[width=.40\textwidth]{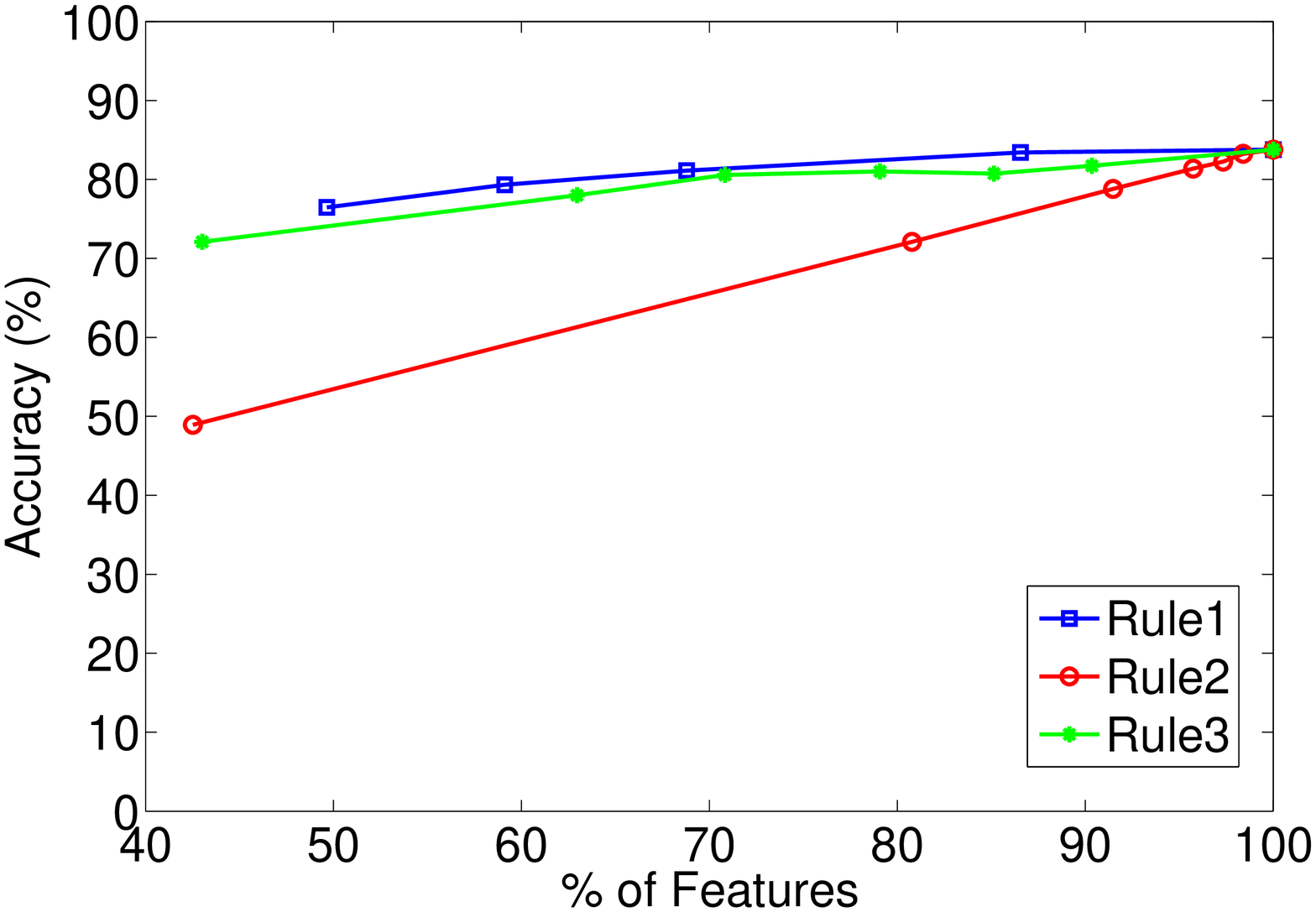}
\end{tabular}
\caption{Accuracy (recall) vs. \%  of features used, City Center dataset}
\label{fig:rules2}

\end{figure}

We used different stopping rules in our experiments and compared the efficiency of each of them when applied to the localization problem. Three rules have been used that are explained below:
\begin{itemize}
  \item[--] Rule 1: computes the distance between the peak and the average of the similarity histogram (i.e., the histogram that shows the similarity of database images to the query). Once the distance is above some threshold, it stops the search: Stop if $|\max(hist)-\mbox{mean}(hist)| > T$ and return the image corresponding to the peak. \\

  \item[--] Rule 2: computes the relative distance between the peak and the average of the histogram. Once the distance is above some threshold it stops the search: Stop if $|\max(h)-\mbox{mean}(h)|/\mbox{mean}(h) > T$ and return the image corresponding to the peak. \\

  \item[--] Rule 3: keeps track of the peak of the histogram. If the peak does not change after processing $T$ features, the image corresponding to the peak is returned as the nearest neighbor to query.

\end{itemize}

\begin{figure}[t]
\centering%
 \tabcolsep 1pt
\begin{tabular}{ccc}
 \includegraphics[width=.35\textwidth]{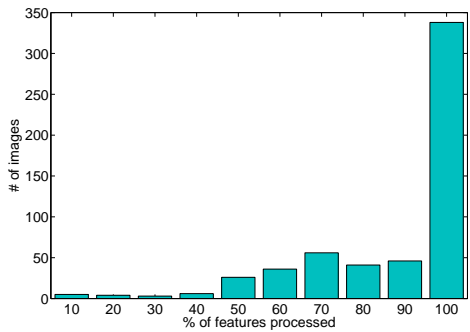} \\

\end{tabular}
\caption{Our experiment on City Center dataset with our stopping method that shows only $60\%$ of images need to process $90\%$\textendash$100\%$ of their features to get the accuracy of BoW.}
\label{fig:feat_im}

\end{figure}

\vspace{.3cm}
The experimental results shown in Tables \ref{exp1:city1} to \ref{exp1:newcol2} have been performed by Rule 1. Figures~\ref{fig:rules} and \ref{fig:rules2} show the result of applying different rules to the BoW search on the New College (right-side sequence) and City Center (right-side sequence) datasets, respectively. The graphs have been generated by varying the stopping thresholds. As can be seen, stopping rules \#1 and \#3 generate better results. By reducing the number of processed features to half the accuracy only decreases slightly.

\vspace{.4cm}

Figure~\ref{fig:feat_im} shows another experiment that validates our proposed stopping method. Figure shows the relation between the number of images and the percentage of features they need to process to obtain the same result as that of the original BoW. The experiment has been done on City Center dataset (left-side sequence) with the stopping rule \#1 used. We set the threshold to $0.29$ to get exactly the same accuracy of original BoW. As can be seen, only $338$ images need to process $90\%$\textendash$100\%$ of their features, another $223$ only need to vector-quantize smaller percentages of features.

\section{CONCLUSION}
\label{sec:conclud}
We observed that the computational requirements of the BoW method can be significantly reduced by allocating less computations to easier search queries. Deciding when to terminate the search is treated as a stopping problem. We proposed several stopping rules and showed their effectiveness on an appearance-based localization problem.


\end{document}